%% file: main.tex
\documentclass[11pt,letterpaper]{article}

\providecommand{\TRPrimaryHex}{CF4E08}
\providecommand{\TRInkHex}{252525}
\usepackage{liveaigc_report}

\input{metadata}

\begin{document}
\maketitle

\input{sections/abstract}

\FloatBarrier
\input{sections/introduction}

\FloatBarrier
\input{sections/related_work}

\FloatBarrier
\input{sections/training_data}

\FloatBarrier
\input{sections/method}

\FloatBarrier
\input{sections/experiments}

\FloatBarrier
\input{sections/deployment}

\FloatBarrier
\input{sections/limitations}

\FloatBarrier
\input{sections/conclusion}

\FloatBarrier
\printcontributions
\printacknowledgments

\FloatBarrier
\bibliographystyle{unsrtnat}
\bibliography{references}

\end{document}

%% file: metadata.tex
\title{TBDub: Production-Oriented Visual Dubbing}
\reportauthors{Bihan Li\textsuperscript{*}, Xinyang Li\textsuperscript{*}, Zeran Xu, Meiguang Jin\textsuperscript{\textdagger}, and Junfeng Ma}
\authornote{\textsuperscript{*}Equal contribution.\quad \textsuperscript{\textdagger}Corresponding author.}
\affiliation{TaoLive AIGC, Taobao \& Tmall Group of Alibaba}
\authoremail{\texttt{\{libihan.lbh, lyon.lxy, xuzeran.xzr, meiguang.jmg\}@taobao.com}}
\teamname{}

\shorttitle{TBDub}
\reportlabel{Technical Report}
\reportdate{September 2026}

\leftlogo[height=0.9cm]{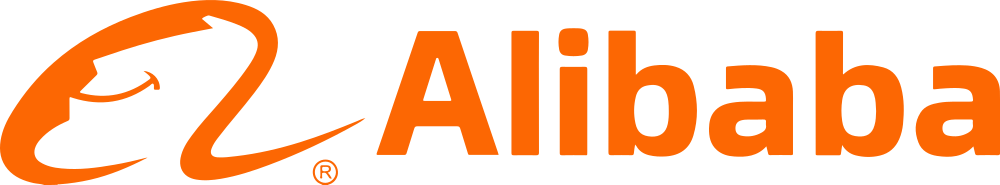}
\rightlogo[
  height=0.9cm,
  trim=436bp 638bp 306bp 480bp,
  clip
]{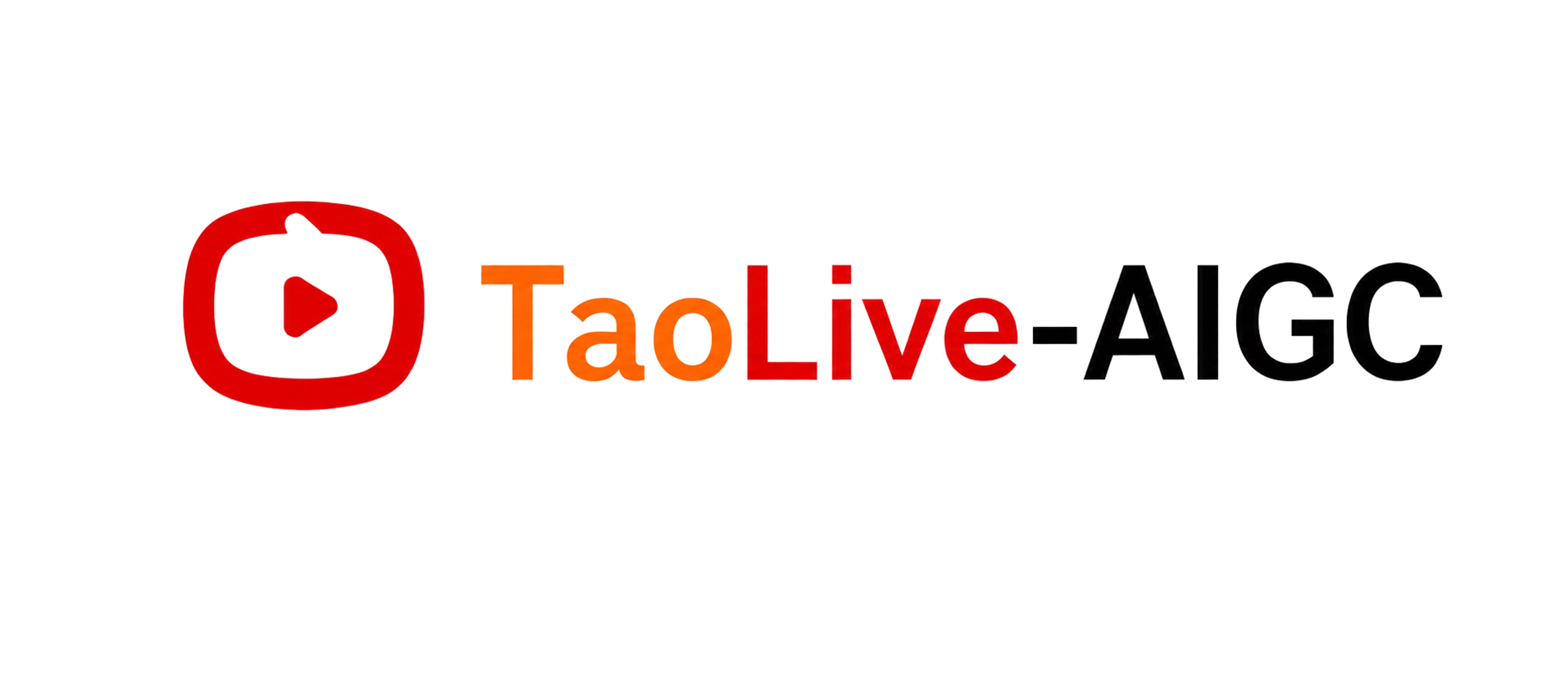}

\projectleaders{}
\corecontributors{}
\contributors{}
\equalcontrib{}

\acknowledgments{}

\hypersetup{
  pdftitle={TBDub: Production-Oriented Visual Dubbing},
  pdfauthor={Bihan Li, Xinyang Li, Zeran Xu, Meiguang Jin, and Junfeng Ma},
  pdfsubject={Task-adaptive post-training and few-step distribution matching distillation of X-Dub},
  pdfkeywords={visual dubbing, lip synchronization, post-training, video diffusion, distribution matching distillation}
}


%% file: sections/abstract.tex
\begin{abstract}
Visual dubbing must synchronize mouth motion with replacement speech while preserving identity, appearance, and temporal consistency. Although X-Dub provides a strong mask-free video-editing baseline, its application to livestream and generated-video content reveals limitations in production-domain robustness, temporal and motion stability, identity and oral-detail preservation, and inference efficiency. We present \textbf{TBDub}, a production-oriented extension of X-Dub that combines task-adaptive post-training with task-aware few-step distillation. Post-training adapts the video DiT using production-domain data, production-specific conditioning and filtering, and enhanced audio features to obtain a 30-step Teacher. Distillation adapts DMD/DMD2 to conditional video editing and compresses the Teacher into a two-step Student. On 38 TalkVid clips, the Teacher improves all eight reported reconstruction, perceptual, identity, and synchronization metrics over X-Dub. In the MOS evaluation, it improves lip-sync consistency, identity consistency, and visual quality over X-Dub by 0.14, 0.95, and 0.90 points, while the Student achieves the highest lip-sync and visual-quality scores and remains close to the Teacher in identity consistency. In paired end-to-end generation timing from the first VAE encode through the final VAE decode on a single NVIDIA H20 GPU at $512\times512$, the Student reaches 7.13 effective FPS and reduces total latency by $13.93\times$; the DiT stage alone is accelerated by $42.49\times$. The Student largely retains the Teacher's generation quality and audiovisual synchronization. The code is available on GitHub at \url{https://github.com/TaoLiveAIGC/TBDub}, and the 30-step Teacher and two-step Student weights are available on Hugging Face at \url{https://huggingface.co/TaoLiveAIGC/TBDub}.
\end{abstract}

%% file: sections/introduction.tex
\section{Introduction}
\label{sec:introduction}

Visual dubbing changes the speech content of an existing video by regenerating audio-correlated facial motion while preserving the subject, camera motion, expression, illumination, occlusions, and background. It is therefore more constrained than unconstrained talking-head generation: the desired edit must be large enough to express new phonetic content, yet small enough to remain visually indistinguishable from the source outside speech-related regions. Errors that appear minor in a single frame---a changed tooth shape, unstable lip color, or a small identity shift---often become conspicuous in video.

Recent diffusion-based systems have substantially improved this trade-off. LatentSync~\citep{li2024latentsync} and KeySync~\citep{bigata2025keysync} use strong latent generative priors for localized lip synchronization. X-Dub~\citep{he2025xdub} goes further by formulating the task as mask-free, reference-conditioned video editing and constructing training pairs through generative self-bootstrapping. Its full visual context and video diffusion prior make X-Dub a capable starting point for our production setting. TBDub builds on this architecture with production-oriented post-training and few-step distillation.

Our production inputs are broader than clean research data. They include controlled green-screen footage, compressed livestream recordings, and videos produced by upstream speech-driven generation systems. In these domains, we observe six recurring classes of failure. First, a full-face editable generator can change identity-sensitive texture rather than only articulation. Second, small faces receive weak effective supervision and lose high-frequency lip and tooth structure. Third, defects already present in an upstream mouth---blurred or fused teeth, green color spill, and lip--tooth adhesion---can be copied through the reference branch. Fourth, local illumination, color, and optical-flow discontinuities can cause flicker, trajectory drift, and boundary jitter. Fifth, robustness remains insufficient for occlusions, large poses, rapid motion, shot cuts, and videos generated by upstream AIGC systems. Finally, abrupt acoustic and alignment changes at the junction between two speech segments can disrupt mouth motion and, in severe cases, cause segment-generation failure. Even when quality is acceptable, the original 30-step sampler remains too expensive for practical throughput.

TBDub addresses these issues through two connected workstreams. \emph{Task-adaptive post-training} extends X-Dub's bootstrapping, filtering, and relighting pipeline with production-domain data, audiovisual calibration, production-specific failure rules, asymmetric condition degradation, an identity-matched clean oral prior, and multi-layer HuBERT speech features. The resulting 30-step model serves as the production Teacher. \emph{Task-aware few-step distillation} combines DMD/DMD2 distribution matching with X-Dub's native three-route dynamic CFG, then adapts it to conditional video editing through a differentiable two-step trajectory, consistent condition handling, and temporary region-selective reconstruction supervision.

The contributions documented in this report are:
\begin{itemize}
  \item we adapt X-Dub to production data and task-specific failure modes through calibrated production-domain data curation, asymmetric lighting and oral degradation with an identity-matched clean oral prior, and full-DiT post-training with motion dropout, spatial face/mouth weighting, later-frame and cold-start temporal weighting, and multi-layer HuBERT conditioning;
  \item we extend DMD/DMD2 to two-step visual dubbing by constructing the real score from X-Dub's three-route dynamically guided Teacher prediction, differentiating through the complete two-step video trajectory, and using staged ground-truth warmup with region-selective supervision to stabilize identity and oral detail;
  \item experiments show stronger qualitative robustness on complex production inputs, improved visual quality and identity preservation, and well-preserved audiovisual synchronization. Compressing sampling from 30 steps to 2, the Student reaches 7.13 FPS and achieves $13.93\times$ end-to-end and $42.49\times$ DiT speedups over the Teacher while largely retaining its quality.
\end{itemize}

\paragraph{Open-source release}
The implementation is publicly available on GitHub at \url{https://github.com/TaoLiveAIGC/TBDub}. The undistilled 30-step Teacher and distilled two-step Student weights are available on Hugging Face at \url{https://huggingface.co/TaoLiveAIGC/TBDub}.

%% file: sections/related_work.tex
\section{X-Dub Baseline and Production Gaps}
\label{sec:baseline-gaps}

\subsection{Task Definition}

Let $\mathbf{V}=\{\mathbf{I}_t\}_{t=1}^{T}$ be a source video and $\mathbf{A}$ a replacement speech signal. Visual dubbing produces
\begin{equation}
  \widehat{\mathbf{V}}=\mathcal{F}_{\theta}(\mathbf{V},\mathbf{A}),
  \label{eq:visual-dubbing}
\end{equation}
such that mouth motion follows the phonetic content and timing of $\mathbf{A}$ while non-speech content remains consistent with $\mathbf{V}$. Unlike self-reconstruction, replacement audio does not define a unique frame-aligned ground truth. A useful system must therefore balance three objectives that can conflict: audiovisual response, visual preservation, and temporal naturalness.

Earlier systems commonly redraw an explicitly selected lower-face area. Wav2Lip~\citep{prajwal2020wav2lip}, VideoReTalking~\citep{cheng2022videoretalking}, DINet~\citep{zhang2023dinet}, and IP-LAP~\citep{zhong2023iplap} established effective synchronization, deformation, and identity-prior mechanisms. Diffusion and latent-space systems such as Diff2Lip, DiffDub, LatentSync, MuseTalk, and SayAnything have progressively improved local realism and audio-conditioned mouth synthesis~\citep{mukhopadhyay2023diff2lip,liu2023diffdub,li2024latentsync,zhang2024musetalk,ma2025sayanything}. X-Dub instead exposes complete reference-video context to a video diffusion model and removes an explicit inference-time mouth mask~\citep{he2025xdub}. This formulation is attractive for difficult poses and occlusions because the model can reason about the complete face and scene, but it also gives the generator freedom to modify content that should remain unchanged.

\subsection{Inherited X-Dub Structure}

We use the public X-Dub system as our baseline. X-Dub inherits a Wan-style latent video generation stack~\citep{wan2025wan}: a causal video autoencoder maps clips to a compressed spatiotemporal latent space, and a Diffusion Transformer (DiT) predicts the denoising or flow field over video tokens. Source-reference latents and target latents are modeled jointly, time-aligned speech features enter through conditional attention, and classifier-free guidance (CFG)~\citep{ho2022cfg} controls the strength of reference preservation and audio response. Motion context can be carried between fixed-length segments for longer sequences. The original inference recipe uses 30 denoising steps.

Table~\ref{tab:contribution-boundary} makes the ownership boundary explicit. This distinction is important because the strongest architectural properties of TBDub---mask-free editing, full-reference context, and the underlying video DiT---come from X-Dub and Wan. Our work starts from the production gaps of this baseline.

\begin{table}[!htbp]
  \centering
  \caption{Technical boundary between inherited mechanisms and TBDub's task-specific adaptations.}
  \label{tab:contribution-boundary}
  \small
  \begin{tabularx}{\linewidth}{@{}>{\raggedright\arraybackslash}p{0.18\linewidth}>{\raggedright\arraybackslash}X>{\raggedright\arraybackslash}X@{}}
    \toprule
    \tablehead{Layer} & \tablehead{Inherited or adopted mechanisms} & \tablehead{TBDub task adaptation} \\
    \midrule
    Base model & Causal video VAE, latent video DiT, reference-conditioned mask-free editing & Full-DiT post-training with motion dropout and spatial/temporal weighted flow matching \\
    Conditions & X-Dub reference-video, audio-conditioning, motion-context, and CFG interfaces & Replacement of the original Whisper/wav2vec~2.0 audio path with an English-pretrained HuBERT-large frontend, concatenating layers 9--12 with a learned projection and segment-level temporal alignment \\
    Data & X-Dub generative bootstrapping, identity/articulation/visual-quality filtering, and relighting & Production-domain data ingestion, audiovisual calibration and production-specific failure rules, $3{:}1$ mixture, asymmetric condition degradation, and an identity-matched clean oral prior \\
    Distillation and sampling & X-Dub three-route dynamic CFG; DMD score difference and online FakeScore; DMD2 two-timescale updates and multi-step generation & Guided Teacher as the DMD real-score, differentiable two-step generation, condition-state handling, temporary region-selective supervision, and causal first-latent handling \\
    \bottomrule
  \end{tabularx}
\end{table}

\subsection{Production Failure Modes}

\paragraph{Identity and non-target drift}
Because the whole face is editable, stronger generative guidance may change skin texture, face shape, eye appearance, hair, or local illumination together with the mouth. The risk grows on long or low-quality inputs and makes a visually plausible frame unsuitable as an edit of a specific person. Reference guidance must therefore be strong enough to preserve the source without suppressing the new articulation.

\paragraph{Small faces and oral detail}
At small face scales, the mouth occupies very few latent tokens. VAE compression and iterative denoising can erase teeth, lip contours, and subtle opening states. Common symptoms include blurred or incomplete teeth, fused upper and lower dentition, teeth visible behind a nominally closed mouth, weak articulation amplitude, and frame-to-frame tooth flicker.

\paragraph{Defect copying from generated references}
In production, the reference may itself be generated by an upstream model. A high-capacity reference branch can faithfully copy undesirable oral artifacts, including green tooth color, smeared texture, and lip--tooth adhesion. This is not solved by general face fidelity alone: training must explicitly separate useful identity appearance from defects that should be repaired.

\paragraph{Illumination and optical-flow discontinuity}
Differences among studio, livestream, and generated-video domains can produce local illumination and color discontinuities, while generated motion may deviate from the source optical-flow trajectory. Even when individual frames appear plausible, these inconsistencies become color or texture flicker, local trajectory drift, and boundary jitter during playback, especially in long videos.

\paragraph{Insufficient robustness to difficult production inputs}
The baseline remains unreliable under occlusion, large poses, rapid motion, shot cuts, and videos produced by upstream AIGC systems. These inputs simultaneously change visible facial regions, motion magnitude, temporal continuity, and reference quality, increasing the risk of incomplete mouth structure, identity drift, unstable edits, or failed segments.

\paragraph{Generation failures at speech-segment junctions}
At the junction between two speech segments, acoustic features and alignment context can change abruptly. If audio feature slicing and video segmentation are inconsistent, the mouth may respond late, jitter, or jump to a discontinuous shape; severe cases can cause the generated segment to fail. Speech conditioning should therefore preserve timing and contextual continuity across the junction.

\paragraph{Inference cost}
Video DiTs are expensive per forward pass, and X-Dub requires 30 sampling steps. Reference/audio CFG can further require multiple Teacher evaluations per step. This cost dominates practical throughput and motivates a Student that learns the already-guided Teacher distribution and generates a segment in only two single-route steps.

%% file: sections/training_data.tex
\section{Task-Adaptive Post-Training}
\label{sec:post-training}

The purpose of post-training is not to relearn a general video prior. It is to retain X-Dub's editing capability while exposing the model to the data distributions, conditioning failures, and oral artifacts that matter in production. We organize this stage as a progression from paired-data construction and filtering to domain mixing, targeted conditioning augmentation, oral repair, and speech representation.

\subsection{Production-Domain Extension of the X-Dub Pseudo-Pair Pipeline}

Directly pairing a video with itself creates an undesirable shortcut: the reference already contains the target articulation, so the model can copy the original mouth and ignore the audio. Following X-Dub's generative self-bootstrapping strategy~\citep{he2025xdub}, we construct pseudo-pairs in which the reference and target retain the same subject, pose, motion, illumination, and background but differ in speech-related mouth motion. An existing dubbing model first redraws a source clip using another speech segment. The redubbed result becomes the conditioning reference, while the clean source clip and its aligned audio remain the target. The generated reference is therefore useful for appearance and motion but unreliable as a source of target articulation.

Because generated pairs can contain identity changes, weak mouth edits, crop failures, or temporal artifacts, they pass through a cleaning funnel. Following X-Dub, identity similarity removes samples whose subject changes substantially, a reference--target articulation test rejects pairs that fail to break the copying shortcut, and visual-quality filtering removes low-quality generations. TBDub applies these rules to production-domain data from livestream and green-screen sources and adds audiovisual calibration together with production-specific checks for abnormal crops, strong blur, flicker, mouth jitter, and decoding failures. In this production-domain data pass, 89,462 candidates were processed and 78,161 were retained, corresponding to an effective retention rate of approximately 87.37\%.

\subsection{Controlled Production-Domain Data Mixture}

Our data have two complementary sources. High-definition green-screen recordings provide stable illumination, clean lip and tooth texture, and reliable supervision for identity-specific detail. Livestream footage covers the conditions that dominate practical failures: small faces, compression, occlusion, side views, walking, rapid head motion, shot changes, and complex illumination. Training only on studio footage limits domain coverage, whereas training only on noisy livestream data lowers the quality ceiling of oral supervision. We therefore sample livestream and green-screen data at a $3{:}1$ ratio. This mixture keeps clean studio examples present throughout post-training while emphasizing the difficult production cases that dominate deployment inputs.

\subsection{Conditioning-Only Asymmetric Relighting}

Building on X-Dub's static and dynamic relighting, TBDub changes paired identical transforms into a conditioning-only asymmetric construction. We use person foreground information derived from DWPose~\citep{yang2023dwpose} to augment a fraction of the conditioning clips with relighting, color-temperature changes, and background replacement, while the clean target remains unchanged. This construction asks the network to recover the target appearance from a reference with plausible capture-domain variation rather than reproduce the perturbation. Sampling the augmentation consistently within a clip avoids introducing artificial frame-wise flicker as a training cue.

\subsection{Oral Degradation Modeling}

Generated or heavily compressed reference videos often contain defects that a strong reference branch will otherwise copy. We convert this failure mode into supervised restoration data by degrading only the oral region of the conditioning reference while retaining the original clean target. A DWPose-derived mouth region provides a soft spatial mask, expanded sufficiently to cover lips, visible teeth, and their immediate boundary. The degradation family covers four production-relevant factors:
\begin{itemize}
  \item spatial detail loss through scale reduction, resampling, and blur;
  \item incomplete separation or adhesion between lips and teeth;
  \item fused upper and lower dentition when the mouth is open;
  \item local color contamination, including the green spill observed in upstream generated content.
\end{itemize}
The transformation strength is randomized by grade, but its main parameters and noise realization are shared across the clip. Clip-consistent sampling is important: independent frame-level corruption would manufacture an unrealistic flicker pattern and teach the model to solve an artifact of augmentation rather than the deployment problem. A feathered mask blends the degraded patch back into the conditioning video, while all supervision continues to come from the corresponding clean frames.

\subsection{Identity-Matched Clean Oral Prior}

Reference-image appearance priors can help preserve identity~\citep{zhong2023iplap}. TBDub specializes this idea into an identity-matched clean oral prior. For each identity, candidate frames are ranked using mouth geometry and local visual quality; a high-quality frame with sufficiently visible oral structure is selected from the top candidates and supplied through the appearance-conditioning stream. The dynamic reference still determines pose, motion, and scene context, and the driving audio still determines target articulation. The clean oral prior supplies a stable example of that subject's lip, tooth, and oral texture.

This design creates an intentional division of responsibility: degraded video context provides time-varying geometry, the clean oral prior provides identity-specific appearance, and audio provides phonetic motion. Oral degradation and the clean oral prior are both used in the final post-training configuration. The former creates a supervised restoration task from locally corrupted references to clean targets; the latter contributes identity-consistent high-quality local texture through candidate selection, oral cropping, and appearance-condition injection.

\subsection{Segment-Aligned Multi-Layer HuBERT Conditioning}

The original audio path combined representations from Whisper~\citep{radford2022whisper} and wav2vec~2.0~\citep{baevski2020wav2vec2}. Our post-training path uses a single English-pretrained HuBERT-large encoder~\citep{hsu2021hubert}. We concatenate hidden states from layers 9--12,
\begin{equation}
  \mathbf{H}^{\mathrm{aud}}
  =\operatorname{Concat}\!\left(
  \mathbf{H}^{(9)},\mathbf{H}^{(10)},\mathbf{H}^{(11)},\mathbf{H}^{(12)}
  \right),
  \label{eq:hubert-concat}
\end{equation}
and map them through a learned projection before conditioning the video DiT. The use of adjacent intermediate-to-late layers retains information at several speech abstraction levels without requiring two separately pretrained encoders whose timing and feature distributions must be reconciled.

Audio is extracted with segment context and then indexed by video time, rather than independently encoding hard-cut feature chunks. For each video latent time step, the model attends to a local window of projected HuBERT features around the aligned acoustic position. This preserves phonetic boundaries and nearby coarticulation at pauses and chunk transitions. The 30-step Teacher and two-step Student use the same English HuBERT-large frontend and layer aggregation, and the FakeScore receives the same aligned feature representation during distillation. Keeping this interface identical prevents an audio-feature mismatch from being absorbed into the distillation objective.

\subsection{Post-Trained Teacher}

The Teacher retains a flow-matching objective, following the continuous transport formulations of flow matching and rectified flow~\citep{lipman2023flowmatching,liu2023rectifiedflow}. We reweight its element-wise error along both spatial and temporal dimensions. Face and mouth masks are derived from DWPose landmarks~\citep{yang2023dwpose} and downsampled to the VAE latent resolution, increasing supervision on the face and particularly on the mouth. Later frames in a target sequence of length $F$ receive linearly increasing weights, and the cold-start first frame receives an additional weight when no motion context is provided.

Let $M_{\mathrm{face}}(i)$ and $M_{\mathrm{mouth}}(i)$ indicate whether latent token $i$ belongs to the face or mouth; let $k_i$ be its temporal index and $b_m\in\{0,1\}$ indicate the presence of a motion condition. We define
\begin{equation}
  \begin{aligned}
    w_i^{\mathrm{sp}}
      &=1+\lambda_f M_{\mathrm{face}}(i)
        +\lambda_m M_{\mathrm{mouth}}(i),\\
    w_i^{\mathrm{tmp}}
      &=1+\lambda_t\frac{k_i}{F-1},\\
    w_i^{\mathrm{cold}}
      &=1+\lambda_c\mathbb{I}[b_m=0\land k_i=0],\\
    w_i&=w_i^{\mathrm{sp}}w_i^{\mathrm{tmp}}w_i^{\mathrm{cold}},\\
    \mathcal{L}_{\mathrm{wFM}}
      &=\frac{\sum_i w_i
      \left\|\mathbf{v}_{\theta}(i)-\mathbf{v}^{\star}(i)\right\|_2^2}
      {\sum_i w_i}.
  \end{aligned}
  \label{eq:post-training-flow}
\end{equation}
The spatial term emphasizes the face and mouth, the temporal term increases supervision toward later video frames, and the cold-start term strengthens the first latent frame only when previous-segment motion context is absent. Their product is normalized by the total token weight so that the overall loss scale remains stable.

Building on X-Dub's flow-matching objective, generative bootstrapping, base filtering, and relighting, this stage adds production-domain data ingestion and rules, audiovisual calibration, conditioning-only asymmetric degradation, the identity-matched clean oral prior, the $3{:}1$ production-domain data mixture, spatially and temporally weighted flow matching, and the multi-layer HuBERT frontend. The resulting model uses the 30-step sampler and serves as the Teacher for the few-step procedure in Section~\ref{sec:distillation}.

%% file: sections/method.tex
\section{Task-Aware Few-Step Distillation}
\label{sec:distillation}

The post-trained Teacher produces stable results with 30 denoising steps, but its cost is incompatible with the target deployment budget. We therefore compress it into a two-step Student. The central challenge is not merely to match an unconditional video distribution: after a drastic reduction in numerical integration steps, the Student must still preserve the source identity and scene, follow a new audio track, maintain motion context, and retain small high-frequency facial structures.

\subsection{From DMD and DMD2 to Conditional Video Editing}

Distribution Matching Distillation (DMD)~\citep{yin2024dmd} trains a one-step generator by matching its output distribution to that of a pretrained diffusion model. It uses the difference between two score functions: a frozen real-data Teacher estimates the direction toward the target distribution, while a FakeScore model estimates the current Student distribution. Their difference supplies a gradient that moves Student samples toward the Teacher distribution without requiring the Student to reproduce every point on a long denoising trajectory. Classical DMD also uses a regression loss on precomputed Teacher noise--image pairs to stabilize optimization.

Building on DMD's score difference and dynamically updated FakeScore, DMD2~\citep{yin2024dmd2} removes the dependence on a persistent regression term and introduces two-timescale alternating optimization, an instance with five FakeScore updates before each Student update, and multi-step generator training; its general formulation may also include adversarial supervision. Progressive and guided diffusion distillation~\citep{salimans2022progressive,meng2023guideddistillation}, consistency-based few-step methods~\citep{luo2023lcm,wang2023videolcm}, and video-oriented motion or distribution distillation~\citep{zhai2024motionconsistency,zhu2024videodmd} have demonstrated the feasibility of few-step image and video generation. We adopt the mechanisms above from DMD/DMD2, while the optional adversarial branch is disabled in the final training configuration.

Applying DMD2 to X-Dub requires three task-specific adaptations. First, the real score must reproduce the joint output of X-Dub's reference-video CFG and audio CFG. Second, within DMD2's multi-step generation framework, we explicitly unroll a two-step Student trajectory for X-Dub conditional video editing and retain the complete computation graph. Third, conditional video editing requires additional region-selective reconstruction constraints to stabilize identity and high-frequency structure before the FakeScore has adapted.

\subsection{Reusing X-Dub's Native Guidance for the DMD Real Score}

We reproduce the three nested condition routes, the audio-side $\sigma^{1.5}$ schedule, and the reference-side $1.0/0.6/0.3$ piecewise cosine schedule from X-Dub's native inference procedure~\citep{he2025xdub}. The resulting guided score is used as the DMD real score, allowing the two-step Student to learn the conditional generation distribution after X-Dub guidance has been applied.

Let $\mathbf{r}$ denote the reference-video condition, $\mathbf{a}$ the aligned HuBERT features, and $\mathbf{m}$ the shared motion context. At a noisy latent $\mathbf{x}_{\sigma}$, the frozen Teacher is evaluated along three nested condition routes:
\begin{align}
  \mathbf{v}_{0} & =
  \mathbf{v}_{T}(\mathbf{x}_{\sigma},\sigma;\varnothing,\varnothing,\mathbf{m}), \\
  \mathbf{v}_{r} & =
  \mathbf{v}_{T}(\mathbf{x}_{\sigma},\sigma;\mathbf{r},\varnothing,\mathbf{m}), \\
  \mathbf{v}_{ra} & =
  \mathbf{v}_{T}(\mathbf{x}_{\sigma},\sigma;\mathbf{r},\mathbf{a},\mathbf{m}).
  \label{eq:three-cfg-routes}
\end{align}
The guided Teacher velocity is
\begin{equation}
  \widetilde{\mathbf{v}}_{T}
  =\mathbf{v}_{0}
  +s_r(\sigma)(\mathbf{v}_{r}-\mathbf{v}_{0})
  +s_a(\sigma)(\mathbf{v}_{ra}-\mathbf{v}_{r}).
  \label{eq:three-route-cfg}
\end{equation}
The first difference isolates reference preservation, while the second isolates the incremental effect of audio given the reference. This nested structure prevents audio guidance from being conflated with the much larger appearance and scene difference between conditional and unconditional generation.

Following X-Dub's native inference recipe, both guidance scales vary with the noise level. Audio guidance follows $s_a(\sigma)=s_a^{0}\sigma^{1.5}$, concentrating strong articulation control at high noise and reducing it during detail refinement. Reference guidance uses a smooth piecewise schedule whose high-, middle-, and low-noise plateaus are respectively $1.0$, $0.6$, and $0.3$ times its configured scale, with cosine transitions between regimes.

Beyond reproducing the native three-route guidance and schedules, we specially handle the first temporal latent. In the Wan causal VAE, the first latent group represents the first pixel frame differently from subsequent latent groups. Applying the full reference schedule at this position can over-guide the first frame and cause blur or texture blooming. We therefore fix its reference scale to one, while later latents use $s_r(\sigma)$. The three condition routes and dynamic scales reproduce the guided prediction of the 30-step X-Dub Teacher; the first-latent rule is a concrete implementation adjustment for Wan's causal representation.

\subsection{Differentiable Two-Step Video Generation within DMD2}

The multi-step generation principle comes from DMD2; our concrete implementation realizes it as a two-step conditional video trajectory with the complete computation graph retained. The Student starts from noise $\mathbf{x}^{(0)}$ and follows a two-step sigma schedule $\{\sigma_0,\sigma_1,\sigma_2\}$. At step $k$, it predicts one fully conditioned velocity and applies an Euler update:
\begin{equation}
  \mathbf{x}^{(k+1)}
  =\mathbf{x}^{(k)}
  +(\sigma_{k+1}-\sigma_k)
  \mathbf{v}_{S}\!\left(
  \mathbf{x}^{(k)},\sigma_k;\mathbf{r},\mathbf{a},\mathbf{m}
  \right),
  \qquad k\in\{0,1\}.
  \label{eq:student-euler}
\end{equation}
The computation graph is retained through both updates. Distribution-matching supervision on the final latent therefore reaches both Student evaluations and optimizes the actual two-step trajectory rather than two independent one-step regressions.

The Student receives all positive conditions in a single route. It learns the native X-Dub fused distribution restated in Eq.~\eqref{eq:three-route-cfg}, so deployment no longer executes the Teacher's three branches. Each Student step requires only one conditional DiT forward pass. The Student, Teacher, and FakeScore share the same reference latents, audio features, motion latents, and auxiliary context. This condition-consistency treatment is part of our video-editing adaptation and prevents a condition mismatch from being mistaken for a distribution gap.

\subsection{Conditional Distribution-Matching Objective}

The score-difference and online FakeScore objectives in this section follow DMD/DMD2. TBDub adapts them by using the X-Dub guided Teacher to construct the real score and by consistently passing reference, audio, motion, and auxiliary conditions to the Student, Teacher, and FakeScore. Let $\mathbf{y}=G_{S}(\boldsymbol{\epsilon};\mathbf{c})$ be the final two-step Student latent under the complete condition tuple $\mathbf{c}$. We sample a noise level from the full training-time range and perturb $\mathbf{y}$ to $\mathbf{x}_{\sigma}$. For a flow-velocity model, the predicted clean latent can be written as
\begin{equation}
  \widehat{\mathbf{y}}_{0}
  =\mathbf{x}_{\sigma}
  -\sigma\mathbf{v}(\mathbf{x}_{\sigma},\sigma;\mathbf{c}).
  \label{eq:velocity-to-clean}
\end{equation}
We compute $\widehat{\mathbf{y}}_{T}$ from the guided Teacher and $\widehat{\mathbf{y}}_{F}$ from the online FakeScore. Their residuals relative to the Student output are
\begin{equation}
  \mathbf{p}_{T}=\mathbf{y}-\widehat{\mathbf{y}}_{T},
  \qquad
  \mathbf{p}_{F}=\mathbf{y}-\widehat{\mathbf{y}}_{F}.
  \label{eq:dmd-residuals}
\end{equation}
For each sample, the implementation first computes the mean absolute Teacher residual over all non-batch dimensions,
\begin{equation}
  d_b=\operatorname{mean}_{c,t,h,w}\!\left(
  |\mathbf{p}_{T,b}|
  \right).
  \label{eq:dmd-normalizer}
\end{equation}
The normalized distribution-matching direction and its stop-gradient pseudo-target are
\begin{align}
  \mathbf{g}_{\mathrm{DM},b}
  &=\operatorname{nan\_to\_num}\!\left(
  \frac{\mathbf{p}_{T,b}-\mathbf{p}_{F,b}}{d_b};
  \mathrm{nan}=0,+\infty=0,-\infty=0
  \right), \\
  \mathbf{y}_{\mathrm{tgt}}
  &=\operatorname{sg}(\mathbf{y}-\mathbf{g}_{\mathrm{DM}}), \\
  \mathcal{L}_{\mathrm{DM}}
  &=\frac{1}{2}\left\|\mathbf{y}-\mathbf{y}_{\mathrm{tgt}}\right\|_2^2.
  \label{eq:dmd-loss}
\end{align}
No epsilon is added to $d_b$, and the quotient is not clipped. PyTorch's \texttt{nan\_to\_num} maps any resulting NaN, $+\infty$, or $-\infty$ value to zero, matching the implementation. Known motion-context positions are removed from this loss because they are fixed conditioning states rather than samples the Student should learn to regenerate. The FakeScore is trained on detached current Student outputs with the standard conditional flow-matching target and is updated separately from the Student. Sampling the complete time range, including the low-noise tail, retains supervision for fine teeth, lips, eyes, hair, and identity texture.

\subsection{Ground-Truth Warmup and Region-Selective Supervision}

The Student and FakeScore are initialized from the Teacher. At the beginning of distillation, their predictions are consequently very similar, while the FakeScore has not yet tracked the Student's emerging two-step distribution. The score difference can then be weak or unstable, permitting identity drift, hair artifacts, or structural collapse. We add ground-truth latent MSE during a finite warmup period to first establish a basic two-step mapping. A global reconstruction term, however, is poorly suited to visual dubbing: in eye and lower-face regions, blinking and audio-driven articulation admit multiple valid outcomes, so global matching would force the model toward a single recorded frame.

Let $M_{\mathrm{eye}}$, $M_{\mathrm{lower}}$, and $M_{\mathrm{lip}}$ be masks at latent resolution, with the lip area contained in the lower-face mask. We use the selective weight
\begin{equation}
  W=\operatorname{clip}\!\left(
  1-M_{\mathrm{eye}}-M_{\mathrm{lower}}
  +\lambda_{\mathrm{lip}}M_{\mathrm{lip}},
  0,1\right),
  \qquad 0<\lambda_{\mathrm{lip}}<1,
  \label{eq:region-weight}
\end{equation}
and define
\begin{equation}
  \mathcal{L}_{\mathrm{GT}}
  =\frac{\sum_i W_i\|\mathbf{y}_i-\mathbf{y}^{\star}_i\|_2^2}
  {\max\!\left(\sum_i W_i,1\right)}.
  \label{eq:region-gt-mse}
\end{equation}
In Eq.~\eqref{eq:region-gt-mse}, $i$ ranges over batch, channel, time, and spatial elements, and $W$ is broadcast over the batch and channel dimensions. The denominator floor of one corresponds to the implementation's \texttt{clamp\_min(1.0)} operation. The eyes and most of the lower face are excluded from direct frame matching; a small lip weight restores enough oral supervision to discourage collapse and severe blur without requiring the Student to copy a unique mouth trajectory. At optimization step $n$, the Student objective is
\begin{equation}
  \mathcal{L}_{S}(n)
  =\mathcal{L}_{\mathrm{DM}}
  +\lambda_{\mathrm{GT}}(n)\mathcal{L}_{\mathrm{GT}}.
  \label{eq:student-objective}
\end{equation}
We set $\lambda_{\mathrm{GT}}(n)=2.0$ for the first 2,000 Student optimization steps and zero thereafter, and use $\lambda_{\mathrm{lip}}=0.1$ in Eq.~\eqref{eq:region-weight}. This term stabilizes entry into distribution matching and is not an inference-time constraint. Unlike DMD's persistent regression to precomputed Teacher samples, our warmup directly uses the real target latent available in the video-editing training pair, applies the region mask in Eq.~\eqref{eq:region-weight}, and is retired once the Student and FakeScore have separated sufficiently for distribution matching to dominate.

\subsection{Optimization Configuration}

The final distillation configuration uses AdamW for both trainable networks. The Student learning rate is $2\times10^{-6}$ and the FakeScore learning rate is $1\times10^{-6}$, with weight decay $0.01$. Following DMD2's two-timescale update scheme and $5{:}1$ update ratio, we perform five FakeScore updates before each Student update; successive updates consume different minibatches rather than repeatedly reusing one sample. The distribution-matching loss has unit weight, gradient accumulation is one, and the retained training schedule runs for ten epochs. The score-training noise schedule uses shift $5.0$, while the two-step Student sampling grid uses shift $1.0$. Optional adversarial, perceptual, identity, and temporal objectives are disabled in this configuration. The CFG combination in the real-score branch is computed in FP32 to reduce additional rounding error when subtracting two similar BF16 predictions.

During training, we smooth the Student weights with an FP32 exponential moving average (EMA) using decay $\beta=0.97$. DMD2 alternates Student and FakeScore optimization while the Student distribution estimated by the FakeScore continues to change, so the raw Student parameters can fluctuate between adjacent optimization steps. A decay of $0.97$ corresponds to an effective weighted window of approximately 33 Student updates; averaging over this window reduces incidental gradient noise and checkpoint-to-checkpoint jitter. The smoothed weights also improve the stability of two-step inference, reducing occasional identity drift, texture flicker, and local structural artifacts. Finally, although the trainable parameters use BF16, the EMA shadow parameters are stored in FP32. The small increment $0.03(\theta_t-\bar{\theta}_{t-1})$ can therefore accumulate at higher precision instead of being lost to BF16 rounding. The two-step Student reported in Table~\ref{tab:horizontal-comparison} uses the EMA checkpoint rather than the raw weights from the final optimization step.

\subsection{Task Adaptations beyond DMD/DMD2}

DMD's score difference and dynamic FakeScore, DMD2's two-timescale alternating optimization and multi-step generation framework, and X-Dub's three nested condition routes and dynamic guidance scales jointly form the foundation of TBDub's few-step distillation.

TBDub connects X-Dub's native guided score to the DMD real score; supplies consistent reference, audio, motion, and auxiliary conditions to the Student, Teacher, and FakeScore; preserves gradients through the complete two-step video trajectory; excludes known motion context from the generation loss; and adds staged region-selective supervision on real data together with causal first-latent handling. Full-timestep sampling, FP32 CFG arithmetic, and FP32 EMA respectively cover the complete noise range, stabilize guided-score computation, and smooth the Student weights.

%% file: sections/experiments.tex
\section{Experiments}
\label{sec:experiments}

We evaluate TBDub through complementary objective and subjective analyses. The objective comparison places the post-trained Teacher and two-step Student alongside representative visual-dubbing systems and characterizes the quality--efficiency trade-off introduced by distillation. A subjective mean opinion score (MOS) study further evaluates lip-sync consistency, identity consistency, and visual quality.

\subsection{Evaluation Protocol}

The horizontal evaluation uses an internal test set of 38 background clips randomly selected from TalkVid~\citep{chen2025talkvid}, with representative variation in language, pose, occlusion, and rapid motion. The self-driven and cross-driven evaluations use the same 38 background videos. In the self-driven setting, each video is paired with its matching audio and therefore has frame-aligned ground truth for full-reference metrics. In the cross-driven setting, the audio is replaced and no unique frame-aligned visual target exists; we therefore assess the outputs using qualitative inspection, no-reference synchronization metrics, Response Mouth PSNR, and subjective MOS.

The qualitative protocol further organizes cases into six categories. Multilingual self-driven reconstruction covers Chinese, English, Korean, Russian, and Japanese. AIGC-domain evaluation redraws MiniMax H3 generated videos using their accompanying generated audio. The cross-driven cases cover large poses, hand/microphone/object occlusions, rapid head motion, and unconstrained in-the-wild videos. All compared methods receive identical source videos and audio tracks under matched resolution, frame-rate, and encoding settings. The MOS study uses the same 38 TalkVid videos: six anonymized method outputs are independently randomized for each rater and video, and three raters score lip-sync consistency, identity consistency, and visual quality.

Each method follows the preprocessing required by its own algorithm. For objective metric computation in Table~\ref{tab:horizontal-comparison}, each source video is independently cropped to $512\times512$ around its median face box and encoded as a lossless FFV1 video rather than composited back into the full frame. All generated outputs then pass through the same evaluation-cropping procedure, and mouth metrics are computed on a fixed region of interest within the crop. PSNR, SSIM, LPIPS, and frame-wise CSIM use method-specific frame maps and ground-truth alignment to avoid artifacts caused by different timeline conventions. FID is computed over the complete batch of frames. All other metrics are first computed per video and then macro-averaged within each method, giving every video equal weight.

The comparison includes X-Dub~\citep{he2025xdub}, LipForcing~\citep{cho2026lipforcing}, KeySync~\citep{bigata2025keysync}, and LatentSync~\citep{li2024latentsync}. The final two rows report the 30-step post-trained Teacher with non-EMA weights and the two-step Student with EMA weights; both use the English HuBERT-large frontend and the same crop protocol. Reconstruction, identity, and synchronization metrics cover all 38 videos, while ground truth is included only as a synchronization reference. This single table therefore supports both horizontal method comparison and direct analysis of the effect of few-step distillation.

\subsection{Metrics}

Mouth PSNR and SSIM~\citep{wang2004ssim} measure frame-aligned mouth reconstruction, with higher values indicating better reconstruction. LPIPS~\citep{zhang2018lpips} measures perceptual discrepancy on the complete face crop and fixed mouth region, while FID~\citep{heusel2017fid} measures the distributional discrepancy between generated and ground-truth frames; lower values are better for all three. ArcFace cosine similarity (CSIM)~\citep{deng2019arcface} measures identity preservation on frames for which an identity feature is successfully extracted. Neither CSIM nor FID directly measures whether mouth motion follows the driving audio.

Audiovisual alignment is measured with SyncNet-derived LSE-C and LSE-D~\citep{chung2016syncnet}; higher LSE-C and lower LSE-D generally indicate better synchronization. Because synchronization networks can exhibit domain bias and can be overfit by methods that directly use synchronization supervision~\citep{yaman2024avexpert}, these values are treated as diagnostic metrics rather than absolute measures of human perception. Response Mouth PSNR compares the mouth regions of two outputs generated from the same source video but driven by the original and replacement audio, respectively. A lower value indicates a larger mouth-motion change and hence a stronger response to the replacement audio. It is a response-magnitude diagnostic rather than a quality score and therefore has no single preferred direction.

For subjective evaluation, lip-sync MOS measures agreement between mouth motion and speech content and rhythm; identity MOS measures agreement with the reference subject; and visual-quality MOS covers naturalness, temporal stability, deformation, flicker, and other visible artifacts. Each dimension uses an integer scale from 0 to 5. The qualitative comparison examines the same properties directly in video, with additional attention to lip and tooth clarity, preservation of foreground occluders, pose-consistent mouth geometry, local illumination, tracking errors, and identity drift. Full-reference reconstruction metrics primarily describe the self-driven setting, while cross-driven dubbing has no unique frame-aligned target; objective metrics and MOS are therefore interpreted together with the six-category video comparison rather than as substitutes for it.

\subsection{Objective Comparison}

Table~\ref{tab:horizontal-comparison} summarizes reconstruction, perceptual quality, identity, audiovisual synchronization, and audio-response metrics for representative visual-dubbing methods, the post-trained Teacher, and the two-step Student. The Student uses the final FP32 EMA checkpoint and was trained with 2,000 steps of real-data MSE warmup and region-selective supervision.

\begin{table}[!htbp]
  \centering
  \caption{Objective comparison on 38 TalkVid clips. Bold and underline indicate the best and second-best results, respectively, in each metric column with a stated preference direction; rankings are determined from the unrounded values. Response Mouth PSNR measures response magnitude rather than quality and is not ranked. Ground truth is reported only for synchronization.}
  \label{tab:horizontal-comparison}
  \small
  \setlength{\tabcolsep}{2.8pt}
  \resizebox{\linewidth}{!}{%
  \begin{tabular}{@{}lrrrrrrrrr@{}}
    \toprule
    \tablehead{Method} & \tablehead{Mouth PSNR $\uparrow$} & \tablehead{Mouth SSIM $\uparrow$} & \tablehead{Crop LPIPS $\downarrow$} & \tablehead{Mouth LPIPS $\downarrow$} & \tablehead{FID $\downarrow$} & \tablehead{CSIM $\uparrow$} & \tablehead{LSE-C $\uparrow$} & \tablehead{LSE-D $\downarrow$} & \tablehead{Response Mouth PSNR} \\
    \midrule
    Ground truth              & -- & -- & -- & -- & -- & -- & 6.28 & 8.35 & -- \\
    X-Dub                     & 22.20 & 0.76 & 0.07 & 0.08 & 9.96 & 0.88 & 5.66 & 9.28 & 23.05 \\
    LipForcing                & 24.33 & 0.82 & 0.05 & 0.06 & \underline{4.34} & 0.91 & 6.13 & 8.37 & 29.28 \\
    KeySync                   & 23.23 & 0.82 & \underline{0.05} & 0.08 & 16.19 & 0.86 & 6.97 & \underline{7.82} & 22.61 \\
    LatentSync                & \textbf{31.30} & \textbf{0.95} & \textbf{0.01} & \textbf{0.02} & \textbf{3.70} & \textbf{0.97} & \textbf{7.76} & \textbf{7.00} & 30.98 \\
    \midrule
    TBDub Teacher (30 steps)       & \underline{27.13} & 0.87 & 0.06 & 0.06 & 9.40 & \underline{0.91} & 6.80 & 7.98 & 25.67 \\
    TBDub Student (2 steps, EMA)   & 26.34 & \underline{0.87} & 0.06 & \underline{0.06} & 9.10 & 0.91 & \underline{7.16} & 7.89 & 23.64 \\
    \bottomrule
  \end{tabular}%
  }
\end{table}

Across the 38 videos, all eight reconstruction, perceptual, identity, and SyncNet metrics of the TBDub Teacher move in the preferred direction relative to X-Dub. Response Mouth PSNR increases from 23.05 to 25.67, a gain of approximately $2.62$~dB. Under the definition above, this higher value indicates a smaller difference between mouth outputs driven by the original and replacement audio: the TBDub Teacher preserves the source appearance more strongly, but responds more conservatively to replacement audio. The metric therefore reflects a trade-off between visual preservation and audio response rather than a one-dimensional quality ranking. Table~\ref{tab:mos} complements this diagnostic with human judgments of audiovisual synchronization, identity preservation, and visual quality.

Relative to the Teacher, the two-step Student decreases Mouth PSNR from 27.13 to 26.34, a difference of $0.79$~dB, while CSIM remains 0.91 at the reported precision. Mouth SSIM remains 0.87, and both Crop LPIPS and Mouth LPIPS remain 0.06 at the reported precision; FID decreases from 9.40 to 9.10. For synchronization, LSE-C increases from 6.80 to 7.16 and LSE-D decreases from 7.98 to 7.89. Response Mouth PSNR decreases from 25.67 to 23.64, a difference of $2.03$~dB, indicating a larger mouth response when the driving audio changes. Considering the scope of each metric, these changes do not imply that the Student surpasses the Teacher in every dimension, but they show that two-step generation largely retains the Teacher's visual quality, identity consistency, and audiovisual synchronization.

The broader ranking also shows why reconstruction metrics alone are insufficient. LatentSync obtains the strongest self-reconstruction and FID results while also having the highest Response Mouth PSNR, indicating the most conservative response to replacement audio among the compared outputs. KeySync performs well in Crop LPIPS but less strongly in Mouth LPIPS. Practical visual dubbing should therefore be assessed jointly through reconstruction, response, synchronization, identity, and direct video observation rather than through any single full-reference metric.

\subsection{Subjective MOS Evaluation}

To complement objective metrics with human-perceived quality, we conduct a subjective MOS evaluation on the same 38 TalkVid videos. Each evaluation group contains a ground-truth reference video and outputs from KeySync, LatentSync, LipForcing, the original X-Dub, the TBDub Teacher, and the two-step TBDub Student. Ground truth is shown only to help raters judge identity and overall appearance and is not rated. The interface hides method names, displays the six candidates as A--F, and independently randomizes their order for every rater and video group to reduce method-awareness and position bias.

Raters assign scores along three dimensions. Lip-sync consistency measures how well mouth motion matches the speech content and rhythm. Identity consistency measures agreement with the ground-truth subject in identity, facial structure, and overall appearance. Visual quality measures naturalness, temporal stability, deformation, flicker, and other visible artifacts. All dimensions use an integer scale from 0 to 5, where scores of 0, 1, 2, 3, 4, and 5 denote completely unacceptable, very poor, poor, fair, good, and excellent, respectively. Three raters participate in the evaluation, and the final MOS aggregation contains 114 candidate-level rating records per method. Table~\ref{tab:mos} reports their arithmetic mean for each method and dimension.

\begin{table}[!htbp]
  \centering
  \caption{Subjective MOS results on 38 TalkVid clips. Each method contains 114 rating records. All dimensions range from 0 to 5, with higher values being better. Bold and underline indicate the best and second-best results, respectively, in each MOS column.}
  \label{tab:mos}
  \small
  \setlength{\tabcolsep}{5pt}
  \begin{tabularx}{\linewidth}{@{}Xrrrr@{}}
    \toprule
    \tablehead{Method} & \tablehead{Ratings} & \tablehead{Lip-sync MOS $\uparrow$} & \tablehead{Identity MOS $\uparrow$} & \tablehead{Visual-quality MOS $\uparrow$} \\
    \midrule
    KeySync                  & 114 & 2.75 & 2.40 & 2.33 \\
    LatentSync               & 114 & 1.91 & 2.23 & 1.87 \\
    LipForcing               & 114 & 2.18 & 2.88 & 2.52 \\
    X-Dub                    & 114 & 3.57 & 2.83 & 2.88 \\
    TBDub Teacher (30 steps) & 114 & \underline{3.71} & \textbf{3.78} & \underline{3.78} \\
    TBDub Student (2 steps, EMA) & 114 & \textbf{3.85} & \underline{3.72} & \textbf{3.80} \\
    \bottomrule
  \end{tabularx}
\end{table}

As shown in Table~\ref{tab:mos}, the two-step TBDub Student obtains the highest lip-sync and visual-quality means, at $3.85$ and $3.80$, respectively, while the TBDub Teacher obtains the highest identity-consistency mean of $3.78$. Relative to the original X-Dub, the Teacher improves lip-sync consistency from $3.57$ to $3.71$ ($+0.14$), identity consistency from $2.83$ to $3.78$ ($+0.95$), and visual quality from $2.88$ to $3.78$ ($+0.90$). These results indicate that task-adaptive post-training improves all three subjective dimensions. Relative to the 30-step Teacher, the two-step Student improves lip-sync consistency by $0.14$, decreases identity consistency by $0.06$, and improves visual quality by $0.02$. Thus, the MOS results show that the two-step Student largely retains the Teacher's perceived generation quality while strengthening perceived audiovisual synchronization at substantially lower computational cost.

\subsection{Qualitative Comparison}

We organize qualitative evaluation into six categories. Each example presents the input alongside TBDub, X-Dub, LipForcing, KeySync, and LatentSync. Complete video cases are available on the accompanying project showcase page. Video evidence is more informative than isolated frames for assessing audiovisual alignment, oral detail, and temporal stability.

\paragraph{Multilingual reconstruction}
Chinese, English, Korean, Russian, and Japanese videos are evaluated in the self-driven setting. This test examines adaptation to different phonetic patterns, audiovisual correspondence, and preservation of identity-sensitive properties such as lip color, tooth shape, and facial texture.

\paragraph{AIGC-generated video evaluation}
We use MiniMax H3 to generate paired videos and audio, then redraw each generated video using its accompanying generated audio. This setup emulates a secondary dubbing stage applied after direct AIGC video generation and tests generalization beyond real-video distributions. Because upstream generated mouths may contain blurred teeth, unstable texture, or local color shifts, we focus on articulation accuracy, lip-and-tooth clarity, and whether the editor amplifies pre-existing defects.

\paragraph{Large-pose robustness}
The large-pose subset contains pronounced side views and substantial head rotation. We inspect whether lip placement remains geometrically consistent with head pose and whether profile views produce malformed articulation, boundary displacement, or identity-texture degradation.

\paragraph{Occlusion robustness}
Occlusion cases include hands, microphones, and other objects near the face or mouth. The model must distinguish foreground occluders from the editable oral region, avoid redrawing the occluding object, and maintain continuity before and after the occlusion.

\paragraph{Rapid head-motion robustness}
These sequences contain abrupt turns, jitter, and motion blur. We compare the temporal trajectories of lip position, tooth shape, local illumination, and facial texture, looking for tracking errors, mouth flicker, and identity drift.

\paragraph{In-the-wild evaluation}
Random real-world videos cover diverse identities, capture conditions, compression levels, and background illumination and are driven by mismatched audio. This setting provides a broad assessment of articulation accuracy, identity preservation, and temporal stability under uncontrolled conditions.

Collectively, these categories cover language variation, generated-video domain shift, pose, occlusion, motion, and unconstrained capture conditions. The comparisons show that TBDub maintains stable oral structure and subject appearance over a broad input range while reducing lip flicker, fused teeth, and local identity drift. Complete results are provided on the project page.

%% file: sections/deployment.tex
\section{Inference Efficiency}
\label{sec:deployment}

\subsection{End-to-End Generation Throughput}

We measure end-to-end generation speed on a single NVIDIA H20 GPU at a resolution of $512\times512$. The measured generation boundary starts with the first VAE encoding operation and ends after the final VAE decoding operation. Model loading, HuBERT audio encoding, face detection and cropping, color correction, compositing and paste-back, video encoding, and file writing are excluded. Each method uses its default best-performing inference configuration, and one complete warm-up run is performed before measurement. The non-TBDub baselines in Table~\ref{tab:inference-efficiency} use the same three-video, 15-second-audio benchmark. The final TBDub Teacher and Student values use the matched sample and timing boundaries detailed in the paired profile below.

\begin{table}[!htbp]
  \centering
  \caption{End-to-end generation throughput under the stated VAE-to-VAE boundary on one NVIDIA H20 at $512\times512$. Higher FPS is better; bold and underline indicate the best and second-best results, respectively.}
  \label{tab:inference-efficiency}
  \small
  \begin{tabularx}{0.82\linewidth}{@{}XXr@{}}
    \toprule
    \tablehead{Method} & \tablehead{Model / steps} & \tablehead{FPS $\uparrow$} \\
    \midrule
    KeySync & Default best configuration & 1.05 \\
    LatentSync & Default best configuration & 1.31 \\
    LipForcing & 14B open-source model, full Wan VAE & \underline{2.83} \\
    Original X-Dub & 30 steps & 0.51 \\
    TBDub Teacher & 30 steps & 0.51 \\
    TBDub Student & 2 steps & \textbf{7.13} \\
    \bottomrule
  \end{tabularx}
\end{table}

The TBDub Teacher and the original X-Dub both achieve 0.51 FPS at the reported precision, showing that task-adaptive post-training does not materially change the inference complexity of the 30-step model. The two-step Student reaches 7.13 effective FPS, approximately $13.93\times$, $13.92\times$, $6.79\times$, $5.43\times$, and $2.51\times$ the reported throughput of the TBDub Teacher, original X-Dub, KeySync, LatentSync, and LipForcing, respectively.

\subsection{Paired Stage-Wise Profile}

We further profile the final Teacher and Student on the same 376-valid-frame sample. The pipeline processes it as six 77-frame clips, or 462 computed frames including overlap and padding. The Teacher uses 30 steps with sigma shift 5 and reference/audio guidance scales of 2.5/10; the distilled Student uses two single-route steps with sigma shift 1. CUDA-synchronized wall time includes reference and motion VAE encoding, DiT execution, other in-pipeline computation, and the final VAE decoding operation. It excludes the components listed above. The reported effective FPS uses the 376 valid output frames as its numerator.

\begin{table}[!htbp]
  \centering
  \caption{Paired stage-wise profile for one 376-valid-frame sample. Latency columns are in seconds; percentages show each stage's share of total timed end-to-end generation latency within the VAE-to-VAE boundary. Lower latency and higher effective FPS are better, and \textbf{bold} indicates the best result in each column based on the unrounded values.}
  \label{tab:stage-efficiency}
  \scriptsize
  \setlength{\tabcolsep}{2.7pt}
  \begin{tabularx}{\linewidth}{@{}Xrrrrrr@{}}
    \toprule
    \tablehead{Method} & \tablehead{VAE enc. $\downarrow$} & \tablehead{DiT $\downarrow$} & \tablehead{Other $\downarrow$} & \tablehead{VAE dec. $\downarrow$} & \tablehead{Total $\downarrow$} & \tablehead{Eff. FPS $\uparrow$} \\
    \midrule
    Teacher (30 steps) & 9.29 (1.26\%) & 696.77 (94.83\%) & 5.15 (0.70\%) & 23.54 (3.20\%) & 734.75 & 0.51 \\
    Student (2 steps) & \textbf{9.27} (17.58\%) & \textbf{16.40} (31.09\%) & \textbf{3.55} (6.73\%) & \textbf{23.54} (44.61\%) & \textbf{52.76} & \textbf{7.13} \\
    \bottomrule
  \end{tabularx}
\end{table}

The two-step Student reduces total timed end-to-end generation latency from 734.75 to 52.76 seconds, a $13.93\times$ speedup and a 92.82\% reduction. DiT execution decreases from 696.77 to 16.40 seconds, corresponding to $42.49\times$. The larger DiT gain comes from both reducing 30 sampling steps to two and absorbing X-Dub's three-route dynamic classifier-free guidance: each Teacher step evaluates a merged batch containing the unconditional, reference-only, and reference-plus-audio routes, whereas each Student step uses one fully conditioned route. The ``Other'' column is the residual after subtracting the explicitly timed stages from the synchronized pipeline total.

The profile also exposes the next deployment bottleneck. DiT execution accounts for 94.83\% of Teacher latency but only 31.09\% of Student latency; final VAE decoding becomes the largest Student component at 44.61\%, followed by VAE encoding at 17.58\%. Consequently, fixed codec and pipeline costs limit the measured end-to-end generation speedup to $13.93\times$ even though the DiT itself is accelerated by $42.49\times$.

Together with the quality results in Table~\ref{tab:horizontal-comparison}, these measurements show that the two-step Student substantially improves throughput while incurring only small decreases in Mouth PSNR and CSIM; the remaining perceptual, distribution, and synchronization metrics remain comparable or improve slightly. The Student therefore provides a more favorable quality--efficiency operating point.

%% file: sections/limitations.tex
\section{Limitations}
\label{sec:limitations}

TBDub remains a reference-conditioned video-editing system, so its output quality is bounded by the quality and spatial resolution of the input video. Severe compression, blur, or facial regions represented by very few pixels can limit the recovery of identity-sensitive lip and tooth details. Very low-resolution inputs may therefore require a separate face super-resolution stage, which is outside the scope of the current pipeline.

The evaluation protocol also has intrinsic limits. Full-reference reconstruction metrics are available only in the self-driven setting and cannot fully characterize cross-driven generation, for which no unique frame-aligned visual target exists. Moreover, SyncNet-derived metrics provide useful synchronization diagnostics but have limited and domain-dependent correlation with human perception. Cross-driven quality should therefore be interpreted jointly from objective diagnostics, MOS, and qualitative video comparisons.

%% file: sections/conclusion.tex
\section{Conclusion}
\label{sec:conclusion}

Building on the X-Dub visual-dubbing backbone and the DMD/DMD2 distillation framework, this work develops task-adaptive post-training and two-step distillation for production-oriented visual dubbing. Post-training improves robustness to complex production inputs through production-domain data and rule expansion, audiovisual calibration, asymmetric lighting and oral degradation on conditioning inputs, a clean oral prior, spatial face/mouth and temporal flow-matching weights, a $3{:}1$ production-domain data sampling ratio, and multi-layer HuBERT audio conditioning. Few-step distillation incorporates X-Dub's native guided score into the DMD real score and combines it with a differentiable two-step trajectory, motion-state exclusion, staged region-selective supervision, causal first-latent handling, and FP32 EMA to obtain a Student that generates in two steps.

On 38 TalkVid clips, the TBDub Teacher improves all eight reported reconstruction, perceptual, identity, and SyncNet metrics over X-Dub. The approximately 2.62 dB increase in Response Mouth PSNR reflects a trade-off between visual preservation and response to replacement audio. In the subjective MOS study with 114 rating records per method, the Teacher improves lip-sync consistency, identity consistency, and visual quality over X-Dub by 0.14, 0.95, and 0.90 points, respectively, showing that task-adaptive post-training improves all three subjective dimensions. In the paired TalkVid crop comparison between the Teacher and Student, the two-step Student decreases Mouth PSNR by only 0.79 dB, while CSIM remains 0.91 at the reported precision and the remaining reconstruction, perceptual, distribution, and synchronization metrics remain comparable or improve slightly. In the MOS study, the Student achieves the highest lip-sync and visual-quality means and trails the Teacher in identity consistency by 0.06. Distillation reduces sampling from 30 steps to 2 steps. In the matched speed profile, the Student reaches 7.13 effective FPS and reduces measured end-to-end generation latency from the first VAE encode through the final VAE decode by $13.93\times$ relative to the Teacher, while accelerating the DiT stage by $42.49\times$. The two-step EMA Student largely preserves the Teacher's generation quality and audiovisual synchronization while substantially reducing inference cost, thereby achieving an effective quality--efficiency trade-off.